\documentclass[a4paper,12pt]{article}
\usepackage{geometry,authblk,linguex}
\usepackage{xcolor,footmisc}
\usepackage{tikz}
\usepackage{pslatex,amssymb,amsmath}
\usepackage{graphicx}
\usepackage{url}

\usepackage{natbib}

\usepackage{empheq} %autoload amsmath

\newcommand{\win}{\mathit{Win}}

\newcommand {\csmodels}  {\mbox{$\mid$ \kern -0.5 em $\approx$}}

\newcommand{\game}{\mathcal{G}}

\newcommand{\fh}{\hat{f}}
\newcommand{\play}{\rho}
\newcommand{\vocab}{(V_0\cup V_1)}

\newtheorem{definition}{Definition}

\newtheorem{proposition}{Proposition}
\newtheorem{rem}{Remark}
\newtheorem{cor}{Corollary}

\begin{document}

\title{Interpretive Blindness}

\author{Nicholas Asher\footnote{CNRS, IRIT} and Julie Hunter\footnote{Linagora GSO}}

\maketitle

\begin{abstract}
  We  model here an epistemic bias we call \textit{interpretive blindness} (IB). IB is a special problem for learning from testimony, in which one acquires information only from text or conversation. We show that IB follows from a co-dependence between background beliefs and interpretation in a Bayesian setting and the nature of contemporary testimony.  We argue that a particular characteristic contemporary testimony, \textit{argumentative completeness}, can preclude learning in hierarchical Bayesian settings, even in the presence of constraints that are designed  to promote good epistemic practices. %In a final section, we develop a game theoretic setting in order to establish theoretical complexity results for IB and to investigate strategies to fight it.
\end{abstract}

\section{INTRODUCTION}

%Epistemic biases make possible the acquisition and maintenance of knowledge, as pointed out almost 300 years ago by Hume in {\em A Treatise of Human Nature} %\footnote{Even after the observation of the frequent conjunction of objects,
%we have no reason to draw any inference concerning any object beyond those of which we have had experience. David Hume, {\em A Treatise of Human Nature}, Book I, part 3, Section 12.} 
%and made mathematically precise by \cite{wolpert:macready:1995,wolpert:1996}.  But they can also impede learning.   %Most ML algorithms incorporate a large amount of prior knowledge or biases but often in implicit and hard to discover form \cite{lampinen:vehtari:2001}.  

In this paper, we describe and analyze an as far as we know theoretically un-examined kind of bias, which we call {\em interpretive blindness} (IB).  IB is exemplified by humans (and perhaps soon by sophisticated machine learning algorithms) whose beliefs are guided and shaped by testimony. When learning through testimony---perhaps the primary way that most people acquire information nowadays---% In knowledge from testimony, 
an agent acquires beliefs through conversations with other agents, or from books, newspapers or % TV or web based media outlets and 
social networks, and so on.  %Acquiring beliefs through testimony is now perhaps the main way people  acquire information.  
Typically, such people lack direct access to the phenomena described via that testimony.  Typically too, humans only pay attention to a restricted set of bodies of testimony from a limited number of sources for their information---which makes sense in terms of an agent's limited resources and attention span.  Our paper is about the strategic consequences of opinion diffusion through testimony and the distortions on learning and information that can result.

IB results from this restriction to few sources of testimony and a natural co-dependence between beliefs and interpretation \citep{JOLLI}. Relying on testimony $T$ from a restricted set of sources to update one's beliefs leads to the mutual reinforcement of our confidence in the source and our belief in $T$; this creates a bias that can preclude learning when an agent tries to exploit new data that are incompatible with or simply distinct from $T$. Agents who are interpretively blind will discount any evidence that challenges their beliefs. We use Wolpert's \citeyear{wolpert:2018} extended Bayesian framework to prove our results.

While IB is problematic for a standard Bayesian framework,  it also poses problems for hierarchical Bayesian approaches \citep{gelman:etal:2013}, because testimony from sources on social media like {\em Facebook}, 24/7 media outlets and web interest groups is often {\em argumentatively complete}, a notion we analyze precisely in Section 4; in an argumentatively complete body of testimony $T$, the authors of that testimony can respond to and argue with any doubts raised by other data or arguments in a body $T'$ that might threaten %by the learner 
$T$'s credibility. A skillful climate denier, for example, will always find a way to undercut the most scientifically careful argument. Argumentatively complete testimony thus can undermine higher order constraints and good epistemic practices that should guide first order learning. %learning---and, as a result, teaching and persuading---impossible in certain cases in a higher order Bayesian framework.  

   %Thus, confirmation bias misses the dynamics that characterizes IB. 

%Interpretive bias also affects our success in using language to teach others and achieve our conversational goals because it imprisons a learner or a teacher within a particular bias. We see the effects of IB when two people in a debate or discussion can only talk past each other despite apparent efforts to convince the other of a position, or when a group of people outright reject any claims from certain news sources or experts that are not corroborated by their own sources. In such cases, the background beliefs of one participant or group affect the interpretation of the other's claims in such a way that the latter will be immediately discounted, regardless of how factual and ``objectively'' well-founded the claims are. 

Our paper starts in Section \ref{sec:testimony} by discussing testimony. We then introduce the codependence of belief and interpretation and apply it to the situation of testimony and the sources that support it.  In Section \ref{sec:firstorder} we formally show how IB can result in ordinary Bayesian learning.  Section \ref{sec:higherorder} shows how IB is reinforced in a hierarchical Bayesian learning setting.  Section \ref{sec:games} develops a game theoretic setting to investigate the complexity of IB.  We provide results as to whether it is possible to free agents from interpretive bias in several epistemic settings.

\section{Testimony and sources}\label{subsec:testimony}
IB arises in learning because of a co-dependence %    Linguists typically seek to establish principles about an "objective" or at least communally shared structure and meaning for language.  However, it is undeniable that in interpretation, the process by which an interlocutor or reader recovers meaning from spoken conversation or written text, subjective elements can play an important role.   \cite{JOLLI} formalizes the interactions between an interpreter's background beliefs and his interpretation of a text.  They go further to show that there is a co-dependence 
between beliefs and the interpretation of evidence, in particular its reliability.  In this case, we are talking of the interpretation of written or linguistically conveyed information.  Others have already noted a co-dependence of beliefs and linguistic interpretation\citep{JOLLI}. Consider this  exchange. R: ``Why hasn't the senator commented on the story that he received undisclosed gifts from supporters?'' A: ``The Senator has declared every gift that he has received.'' As \cite{JOLLI} argue, one could interpret A's response either as an answer to R's question or as an evasion and it depends upon one's beliefs about the honesty of the senator.  Those beliefs get confirmed as the exchange and its interpretation continue: the interpretation that relies on the honesty of the Senator gets confirmed as A refuses to engage with R, and this in turn confirms the honesty of the belief; {\em mutatis mutandis} for the other interpretation.

%our beliefs typically affect how we interpret a linguistically given information; they help resolve ambiguous phrases or words one way rather than another, and they can affect how we link  linguistically conveyed propositions together when the author leaves these links underspecified.  On the other hand, the resulting interpretations subsequently serve as evidence upon which we update our beliefs.  The dynamic between belief and interpretation as we read through a text can strengthen our beliefs and simultaneously a particular interpretation of the text\citep{JOLLI}. %---how beliefs affect interpretation, how these interpretations in turn strengthen the beliefs that led to them and how these strengthened beliefs affect further interpretation.  % conversation, an exchange between a reporter and a spokesman for a senator; these interpretations diverge concerning the source's trustworthiness and become contradictory, because of a difference in background beliefs.

An analogous co-dependence occurs with interpretation, belief and learning: in updating our beliefs with new evidence; our beliefs color how we interpret that evidence, in particular how trustworthy we find it.  The updated evidence in turn conditionally updates our beliefs.
%Bayesian updating provides the foundations for the first part of this co-dependence; confirmation bias suggests the second part.  We put the two aspects together and investigate how this co-dependence can iteratively entrench beliefs and close off openness to new evidence---the result of which is IB. % ***is this clear enough??***

Let us look this codependence in learning by testimony.   
  A body of testimony $T$ is a collection of information conveyed by one or more sources like \textit{The New York Times},  {\em Fox News}, {\em CNN}, {\em Facebook}, {\em 4Chan}, a particular individual or set of individuals. The sources may ``promote'' or vouch for $T$ or cast doubt on $T$.  Such bodies are also \textit{dynamic}; they evolve over time as they are updated with new facts and events. In other words,  $T$  comes in ``stages'',  where stages might be defined by times or even conversational turns, and each stage $T_i$ is the body of evidence accumulated up to stage $i$.  $T  = \{T_1, T_2, ... , T_n, ...\}$ is the collection of all the stages of a dynamic body of evidence.  Dynamic bodies of testimony are ubiquitous in our communicative landscape; on-line, 24/7 news sources as well as particular groups on social media provide evolving, updated coverage of new events. Let ${\mathcal T}$ be a collection of bodies of testimony about some phenomenon $P$.  We will assume that $\fh$ does not have independent access to $P$ and uses evidence from bodies of testimony in ${\mathcal T}$ together with background beliefs to update probabilities about hypotheses about $P$, some of which are hypotheses about marginal probabilities of events described in ${\mathcal T}$. 
%Dynamic bodies of evidence are ubiquitous in our communicative landscape; news sources, {\em CNN}, {\em The New York Times}, and, are  dynamic, providing updated coverage of new events.

%Central to our confidence in a body of testimony is its {\em source} or author.  We can only rely on testimony if we find its source credible.  So a crucial parameter in our learning set up is how $\fh$ treats such sources, and that will depend on $\fh$'s background beliefs.  
Learning from testimony $T$ with source $s$ requires a learner $\fh$ to judge $T$ as credible, a judgment that will depend on $s$'s evaluation of $T$ (whether $s$ promotes or challenges $T$), as well as $\fh$'s antecedent hypotheses about $s$.   Let ${\mathcal H}$ be a set of \textit{evaluation hypotheses}, where each $h \in {\mathcal H}$ gives the evaluation of a set ${\mathcal T}$ of bodies of testimony $T$ relative to a source $s$.  $h \in {\mathcal H}$ defines a conditional probability $P(T|h)$ for $T \in {\mathcal T}$, which we will sometimes write  as $h(T)$, where $h(T) = 0$ means $T$ is untrustworthy according to $h$, and $h(T) = 1$ means $T$ is trustworthy ($s$ fully endorses $T$). Following Wolpert's \citeyear{wolpert:2018} extended Bayesian framework, our learner $\fh$ updates his belief in $T$ relative to ${\mathcal H}$.

Our learner $\fh$ will have a probability distribution over his evaluation hypotheses ${\mathcal H}$.  Given the co-dependence of beliefs and evidence, this distribution is updated relative to the stages of $T$ as it develops.  This is intuitive; the testimony $T$ should serve as evidence upon which $\fh$ updates his beliefs.  But the co-dependence tells us that $\fh$ updates his confidence in $T$ via these updated beliefs.  %Given his background beliefs, our learner $\fh$ may interpret testimony $T$ in such a way that $\fh$ affects his beliefs about the trustworthiness of the source of $T$. To this end, we  define an evaluation hypothesis $h: {\mathcal T} \rightarrow [0,1]$, with $h(T) = 0$ for untrustworthy testimony $T \in {\mathcal T}$ and $h(T) = 1$ for trustworthy testimony (full belief).   

%There is a probability distribution over the set ${\mathcal H}$ of $\fh$'s evaluation hypotheses that, given the co-dependence of beliefs and evidence, is updated via new evidence.  This is intuitive: $\fh$ should judge the reliability of an evaluation hypothesis by the content of the testimony it gives rise to and the evidence at his disposal.  Now supposing that testimony $T$ provides $\fh$'s evidence, the more $T$ supports an evaluation hypothesis $s$, the more plausible $s$ becomes, compared to an $s'$ that discounts $T$.  In turn, updated evaluation hypotheses update $\fh$'s beliefs about $T$.  

 %IB arises when $\fh$ puts all of hi subjective probability mass on a set of evaluation hypotheses that count only some bodies of evidence trustworthy.  
 Most if not all of us  acquire new information from a restricted set of bodies of evidence that push a particular point of view. This is reasonable given the balance rational agents need to find between exploiting already acquired data  and gathering more data. In addition, attending to a particular body of evidence can give a sense of community, as has been amply documented in the scholarly literature and the press.  But this trade off can lead to a problem in learning: when we rely on testimony to learn and we restrict the testimony we pay attention to, the confirming evidence for the evaluation hypothesis and what it supports threaten to collapse into one.  % We may have several evaluation hypotheses about the reliability of that evidence because we are unsure or of ``several minds" about what we read, but
 We now turn to see how iterated Bayesian updating in learning from testimony can ultimately lead to a situation where only evaluation hypotheses supporting our restricted evidence are credible and this leads to IB.
%We will also suppose that a learner may have a collection of evaluation hypotheses that may give conflicting advice and over which the learner has a probability distribution.

\section{IB in a first order Bayesian setting}\label{sec:firstorder}

To formalize IB and its consequences, we first present a simple experiment to show how the codependence of interpretation and belief leads to % first prove a folk theorem of 
{\em bias hardening}.  To illustrate, suppose that $\fh$ considers a consistent dynamic body of testimony $T = \{T_1, T_2, ... , T_n, ...\}$ and has two evaluation hypotheses $h_1, h_2$, where the prior probabilities assigned to $h_1$ and $h_2$ by $\fh$ are:
\begin{equation}\label{eq:source}
P(h_1) = .6, \ \ P(h_2) = .4 
\end{equation}
and the evaluation hypotheses assign probabilities to $T$ as it evolves through stages $T_i$ as follows:
%$$h_1(T_1) = .9 = P(T_1|h_1), \ \ h_2(T_1) = .1 = P(T_1|h_2)$$
%and
\begin{equation}\label{eq:stage} P(T_i | h_1) = .8, \ \ P(T_i |h_2) = .2   \mbox{ for all i }  \end{equation}
We can now calculate the probability of $T_1$ using the general rule for marginal probabilities in \ref{sum}. Let ${\mathcal B}$ be $\fh$'s background beliefs; and let the set of all $h_i$, the alternative hypotheses that are consistent with or assigned non-zero conditional probability  relative to ${\mathcal B}$ \citep{lampinen:vehtari:2001,tenenbaum:etal:2006,tenenbaum:etal:2008,tenenbaum:etal:2011}, be the set of evaluation hypotheses $h_i$ (so $\{h_1, h_2\}$, in our example).  %Here our alternative hypotheses are our evaluation hypotheses $h_1$ and $h_2$. 
\begin{equation}\label{sum}
P(x) = \sum_{i = 1}^{i = k} P(x|h_i, {\mathcal B}).P(h_i, {\mathcal B})
\end{equation}
 Then using (\ref{eq:source}), (\ref{eq:stage}), and (\ref{sum}), we have:
\begin{equation}P(T_1) = P(T_1|h_1).P(h_1) + P(T_1|h_2).P(h_2) = .56.
\end{equation}
This is our estimation of our belief in the body of evidence $T$ based on what we have so far.  We will continue to update the probability of $T$ given new stages $T_i$ below by distinguishing prior probabilities $P_{prior}$ and updated probabilities $P_{post}$. Now suppose there is a new conversational turn in $T$, a new stage of evidence $T_2$.  Given our assumptions, $P(T_2|h_1) = .8$, while $P(T_2 |h_2) = .2$, $T_2$ is supported by $h_1$ but not by $h_2$---$h_1$ and $h_2$ are consistent with their roles on $T_1$. Given the dependence of beliefs and interpretation of evidence, $T_2$ also leads us to re-evaluate our evaluation hypotheses by adapting Bayes' formula to our evidential hypotheses:
\begin{equation} \label{update}
P(h_i|T_{n+1}) = \frac{P(T_{n+1}|h_i) P_{prior}(h_i)}{ P_{post}(T_n)}
\end{equation}
Given $T_2$, whose initial probability we set to what the posterior calculated for $T_1$---i.e., $P_{post}(T_1) = P_{prior}(T_2)$, we can update our confidence in $h_1$ as follows: 
\begin{equation} P(h_1|T_2) = \frac{P(T_2 | h_1).P_{prior}(h_1)}{P_{post}(T_1)} \approx .86.
\end{equation}
Thus, we have posterior probabilities for our evaluation hypotheses as well as for stages of bodies of evidence.  The similarly updated probability for $h_2$ now drops to roughly $.14$.  Using the updated values for $h_1$ and $h_2$, we see that $T_2$, which includes $T_1$, is now even more believable: $P_{post}(T_2) = .74$.  Now suppose that a new bit of evidence, $T_3$, is added to $T$. As before, we set $ P_{post}(T_2) = P_{prior}(T_3)$. Given our assumptions about our source functions, $P(T_3|h_1) = .8$, we have $P(h_1|T_3) = .96$, while $P(h_2| T_3) \approx 0.04$, and confidence in $T_3$ is also updated:  $P_{new}(T_3) = .776 \approx .78$.  Updating $h_1$'s probability conditional on new evidence $T_4$ now yields a value of $.989 \approx .99$, while $P(h_2| T_4) = 0.008 \approx 0.01$.
%With these approximations by 
By the time we get to $T_5$, the probability of $h_1$ will have gone to $1$, while $P(h_2) = 0$, and $P(T_5) = .8$.  
In sum, as $n$ increases, the updated probabilities of $h_1$ go to $1$ and $P(T_n) \rightarrow  P(T | h_1)$, that is, to the strength of $h_1$'s support for $T$.

Our co-dependence of belief and evidence suggests a loopy structure (cyclic graph) for updating. However, by exploiting stages, we can disentangle such structures; and efficient approximations are possible in disentangled structures \citep{murphy:etal:2013}.  Proposition 1 below shows a convergence under certain assumptions. Let ${\sf P}_n(h_i)$ be the probability of $h_i$ after conditionalizing on $T_n$ and ${\sf P}_n(T)$ the value of $T$ after n conditional updates as defined above. Generalizing our discussion above, we have:
\begin{proposition}
Suppose testimony $T = \{T_1, T_2, ... , T_n, ...\}$, and suppose $\fh$'s evaluation hypotheses ${\mathcal H}_{\hat{f}}$, % h_{\hat{f}} = \{h_1, h_2 , ... h_i\}$ 
with a probability distribution and with $h_1 \in {\mathcal H}$ such that  $P(h_1) \neq 0$, and $P(T_n | h_1) > .5$ and is monotone increasing for all $n$, while  $P(T_n | h_j) < .5$ and is monotone decreasing for all $n$ and for all $h_j \in {\mathcal H}, h_j \neq h_1$.  Then:%  we have:\\  
$$\mbox{As } n \rightarrow \infty, \  {\sf P}_n(T) \rightarrow  limsup ( P(T_n | h_1)),$$ $${\sf P}_n(h_1) \rightarrow 1 \mbox{ and } {\sf P}_n(h_j) \rightarrow 0 \mbox{ for } j \neq 1$$ \label{codependence}
\end{proposition}   
\noindent Given the calculations above and using standard updating rules for the probabilities $P$ assigned by $\fh$, %{\color{violet} as long as the probability mass on $h_j$ is lower than the probability of $h_1$}.  it is clear that instead of $h_2$ we could have had finitely many
 if $P(T_i | h_1)$ is monotonic increasing with respect to $i$ and $P(T_i | h_j)$ for any $j \neq 1$ is monotonic decreasing, then the updates of $P(T_i)$, $P(h_1 | T_i)$ and $P(h_j | T_i)$ will follow the pattern of our experiment above and converge to the support of $h_1, 1,$ and $0$ respectively. $\Box$

\noindent  %as we have defined it, 
  %More precisely,
%\begin{cor}$P(T_i|s) = 1 \mbox{ for all } i \rightarrow lim_{n \rightarrow \infty} P(s | T_n) = 1$
%\label{codep1} \end{cor}

 We now introduce three important properties of evaluation hypotheses. %, as Proposition \ref{discount} shows.  %%We begin by defining the notion of consistent bodies of testimony.

\begin{definition} An evaluation hypothesis $h$ for a set of bodies of testimony ${\mathcal T}$ is {\sf consistent} iff for $T, T' \in {\mathcal T}$, if $T \cup T'$ is inconsistent, then $P(T|h) = 1 - P(T'|h)$. An evaluation hypothesis $h$ is {\sf probability-wise model complete (PWMC} for $T$  iff: for any putative piece of evidence $\phi$ if for no stage $T_i$ $T_i \models \phi$ ($\phi$ is not predicted or included in any stage of $T$), then $P(\phi| h) = 1 - P(T|h)$.
\end{definition}
\begin{definition}
An evaluation hypothesis $h \in {\mathcal H}$ with a probability distribution $P$ makes $T$ {\sf potentially trustworthy} ($h \models T$), if for all $n$, $P(T_n|h) > .5$ and as $n \rightarrow \infty$, %^1 \in {\mathcal H}, this didn't make sense
$P(T_n|h) \rightarrow 1$.% and $P(h) \neq 0$. 
\end{definition}

\noindent
 We take consistency to be a basic requirement of evaluation hypotheses.   %That is,  probability of a hypothesis as in equation \ref{update}, so too we can update the conditional probability of $T_i$ given an evidential hypothesis
% \begin{equation} \label{update1}
%{\sf P}_{n+1}(T_{n+1}|h_i) = \frac{{\sf P}_n(h_i |T_{n-1}) P_{n}(T_{n+1})}{ P_{n}(h_i)}
%\end{equation}
PWMC hypotheses generalize consistent hypotheses.  But what is their rationale?  As $T$ evolves through its stages, it is natural to assume that $T_{i+1}$ to provide a more complete coverage of the facts than $T_i$.  And as $T$ becomes more complete, an evidential hypothsis may assume elements $\phi$ that are not mentioned in any $T_i$ are in some way incompatible with $T$. The PWMC condition codifies this in terms an operation akin to the operation of negation as failure in Prolog; if $h$ makes $T$ probability wise model complete, then if $T$ doesn't mention $\phi$ then $h$ supports $\neg \phi$ to the extent that $h$ supports $T$.   

In addition, an agent plausibly has, among the many evaluation hypotheses that she countenances, an evaluation hypothesis $h$ for which the conditional probability of $T$ given $h$ increases as $T$ evolves.  Up to now we have taken an evaluation hypothesis to confer a fixed probability on a body of evidence $T$ it supports or discounts.  But the support for $T$ might increase (or decrease) as $T$ gets more extended with more and more stages. A potentially trustworthy evaluation hypothesis is a kind of ``soundness'' or accuracy assumption about a body of evidence.  For an agent who remains wedded to a body of testimony such a soundness assumption seems a rational requirement.  If potential trustworthiness is a soundness requirement then PWMCness is a kind of completeness requirement.  Together they furnish the rational justification for an agent to restrict his learning to a particular body or particular bodies of evidence, something that we've argued most people in fact do and do so with rational justification.

% \ref{codependence}  and takes $T$ to be trustworthy, i.e., $h_1(T)$ = 1,
%Proposition \ref{codependence}  then tells us:
\begin{proposition} \label{discount} Let ${\mathcal T}$ be a set of consistent bodies of testimony and let ${\mathcal H}$ be a set of evaluation hypotheses for ${\mathcal T}$,  %= \{h_1, h_2, ... h_k: T \rightarrow [0, 1] \mbox{ for } T \in {\mathcal T}\}$ 
with $h_1 \in {\mathcal H}$ and $h_1 \models T$, for some $T \in {\mathcal T}$.  Let the priors on $h_i \in {\mathcal H}, h_i \neq h_1$ be as in Proposition \ref{codependence} and let the probabilities of $h_i, h_1 \in {\mathcal H}$ be updated on $T$.  Then: 
\begin{equation*}
    \mbox{As } n \rightarrow \infty, \ {\sf P}_n(T) \rightarrow 1. \tag{1}
\end{equation*}
\noindent
Suppose in addition, $h_1$ is PWMC for $T$ and $T \not\models T'$. Then: 
\begin{equation*}
\mbox{As } n \rightarrow \infty, {\sf P}_n(T') \rightarrow 0  \tag{2}\end{equation*}
%\lim_{n \rightarrow \infty} P(h_1 | T'_n) = 0$$
\end{proposition}
\noindent
 To show (1), note that $P(T_1 | h_1) >.5$ and since as $n \rightarrow \infty$, ${\sf P}_n(h_1) \rightarrow 1$, after a certain point $P(T_n| h_1)$ is monotone increasing.  Then by Proposition \ref{codependence},  ${\sf P}_n(T) \rightarrow P(T_n | h_1)$. Since $h_1$ makes $T$ potentially trustworthy, as $n \rightarrow \infty, {\sf P}_n(T) = 1$. To show (2), suppose %$T \cup T'$ is inconsistent.  Alternatively, suppose 
 $h_1$ is PWMC for $T$. Given that  $T \not\models T'$, $h_i(T'_i) = 1 - h_i(T_i)$ for each $i$, and the expected probability of $T'$ will decrease strictly monotonically over n, as ${\sf P}_n(h_1) \rightarrow 1$.  So as $n \rightarrow \infty,\ {\sf P}_n(T') = 0.$ $\Box$ 

%\vspace{1mm}
\noindent
Note that our agent may have many evaluation hypotheses and the result of Proposition \ref{discount}.  Crucially $\fh$ has updated his beliefs only on $T$.  But this matches our intuitions about what agents actually do.  As long as the co-dependence between background beliefs and bodies of evidence holds and certain bodies of evidence are supported more than others, belief in some bodies of evidence $T \in {\mathcal T}$ will be strengthened, while belief in bodies of evidence in conflict with $T$ or just different from $T$ will be weakened.  Importantly, this can happen {\em merely by } $T_i$ {\em repeating content already in} $T_k$ for $i > k$.  Such repetitions of content are commonplace on social media sites and news sites that broadcast continuously.  In addition, the assumption of a dogmatic evaluation hypothesis is actually mild; it reflects an agent's mistrust of bodies of evidence other than the ones he relies on---a rather common situation. 

Proposition \ref{discount} impacts the marginalization of new data, because if its assumptions are met, as ${\sf P}_n(T') \rightarrow 0$, $\fh$  discounts evidence from $T'$, despite the presence of evaluation hypotheses supporting $T'$.  %Assume that $\fh$ has become aware of some data (description) $x$ such that $T \not\models x$ ($x$ is not predicted or included in $T$) but that $T' \models x$.   %We  write $T \models \neg x$ to mean that $\neg x$ is a consequence of $T$ (and hence that $x$ is inconsistent with $T$.  
\begin{proposition}  Suppose evidence $\phi$ such that  $T' \models \phi$, $T \not\models \phi$ and $T, T'$, and $\fh$'s evaluation hypotheses are as in Proposition \ref{discount} and $\fh$ conforms to Bayesian learning. Then:\\ hspace*{.5in} As $n \rightarrow \infty, \ {\sf P}_n(\phi) \rightarrow 0$. \label{marginal1}
\end{proposition} %As such
%\begin{proposition} \label{unlearning}
%Given $\fh$, ${\mathcal T}, {\mathcal B}, t \in {\mathcal T}$ and $h(t)$ as defined above, $\fh$ will fail to learn correct classifications for 
%all $x \in X$ such that $x$ is indeed a fact but is put forward as a fact only in $t$ for $h(t) = 0$.
%\end{proposition}
%Let  the set of hypotheses $h_i$ in Equation \ref{marginal} be the set $S$ of evaluation hypotheses  that are assigned non $0$ probability relative to ${\mathcal B}$
\noindent
Since $\hat{f}$ conforms to Bayesian learning, the marginal probability for $\phi$ is based on Equation \ref{sum} and the set of hypotheses $h_i$ in Equation \ref{sum} is the set ${\mathcal H}$ that for $\fh$ pronounce on testimony that mentions or asserts $\phi$.  By Proposition \ref{codependence}, as $n \rightarrow \infty$, ${\sf P}_n(h_1) \rightarrow 1$.  By Proposition \ref{discount}, ${\sf P}_n((T'h_1) \rightarrow 0$.  But for all other $h_k$ such that $h_k (T') \neq 0$, by Proposition \ref{codependence} again,  as $n \rightarrow \infty$, ${\sf P}_n(h_k) \rightarrow 0$.    But then ${\sf P}_n(\phi|h_i, {\mathcal B}) \rightarrow 0$ for all relevant $h_i$. Given Equation \ref{sum}, the result follows. $\Box$

\noindent
In this situation, $\fh$ assigns no credence to $\phi$.  The prior beliefs of $\fh$ may so limit the alternative hypotheses $h_i$ such that even an actual fact $\phi$ will have a marginal probability of $0$; $\fh$ will discount $\phi$ completely.%, if her evidence for $x$  that she eventually distrusts completely.  

Now consider general learning in this situation, defined in Wolpert's \citeyear{wolpert:2018} extended Bayesian framework via Bayes's formula below.
\begin{equation}
P(h|x, {\mathcal B}) = \frac{P(x|h, {\mathcal B}).P(h|{\mathcal B})}{\sum_{i = 1}^{i = k} P(x|h_i, {\mathcal B}).P(h_i | {\mathcal B})} \tag{7}
\end{equation}
To learn a hypothesis $h$, $\fh$'s estimation of $h$ at some stage should be closer to the  objective or ideal assignment (posterior) $h_p$ to $h$, than his prior probability for $h$.  Similarly for marginal probabilities: ${\sf P}_n(x)$ should track $x_p$, the posterior of $x$, given a random sampling of $X$.  We consider loss functions ${\mathcal L}({\sf P}_n(h), h_p)$ and  ${\mathcal L}({\sf P}_n(x), x_p)$.  The greater divergence between the ideal posterior probability and the Bayesian subjective estimation of that probability, the worse will be the score for $\fh$'s learning.  We say that $\fh$ cannot learn $h$ if additional evidence does not eventually decrease loss; i.e. we cannot show $ lim_{n \rightarrow \infty} {\mathcal L}({\sf P}_n(h),h_p) < {\mathcal L}({\sf P}_0(h), h_p)$.
\begin{proposition} \label{unlearning}
Suppose $\fh$ is a Bayesian learner with evaluation hypotheses and testimony $T$, $T'$ as in Proposition \ref{discount} and all evidence $e$ confirming $h$ is such that $T' \models e$. Then  $\fh$ is incapable of learning $h$.
\end{proposition}
\noindent Consider $e$ such that $T' \models e$  and e confirms $h$.  So the true posterior $P_p(h|e) > P(h)$, with $P(h)$ the prior on $h$.  Suppose $\fh$'s evaluation hypotheses and probabilities have been updated via $T$ as in Proposition \ref{discount}.  %If $\fh$ assigns $x$ a marginal probability in the the light of equation (\ref{marginal}), then $x$ will be significantly below $\frac{1}{2}$; 
By Proposition \ref{marginal1}, as $n \rightarrow \infty, \ {\sf P}_n(e) \rightarrow 0$.  In the limit, Bayesian learning as specified by equation (4) simply isn't defined when ${\sf P}_n(e) = 0$.  So assuming $e$ is discounted as evidence in updating,
%At any finite stage, because evidence $e$ is increasingly discounted, evidence relevant to $h$ is taken less and less into account.  
we set $P(h | e, T_n) =  P(h | T_n)$.  But this is just $E_0(h)$ or $P(h)$, $\fh$'s prior on $h$.  It follows that as $lim_{n \rightarrow \infty} \ {\mathcal L}({\sf P}_n(h),h_p) \not< {\mathcal L}({\sf P}_0(h), h_p)$. $\Box$

Proposition \ref{unlearning} is a formal statement of IB in a first order setting.  It shows that under certain conditions, $\fh$ will be incapable of learning any hypothesis that involves a dependence on testimony not in  $T$, upon which $\fh$ has formed his beliefs. $\fh$ is interpretively blind to any possibilities outside of $T$.  %This is a type of bias: if we set $\varphi(x)$ to be $x$ {\em is an event or hypothesis the learning of which depends on testimony not in $T$} and set $\delta$ to the proposition that $x$ is learned or assigned a marginal probability, then the following pair of counterfactuals is true in $M_{\hat{f}}$: for some set of literals  $C(x)$, $M_{\hat{f}} \models C(x) \ \cftual \delta$ and $ M_{\hat{f}} \models (\varphi(x) \wedge C(x)) \ \cftual \neg \delta$. This  fits our definition of bias with respect to $\varphi$ of Definition 1.  %In $M_{\hat{f}}$ the conditional probability $P(\delta | \varphi(e))$ is either undefined or simply stipulated to be the prior $P(h)$, as the evidence $e$ in this makes a set of such counterfactuals true, one for each event or hypothesis based on a relevant body of evidence. %In a deterministic setting, discounting amounts to, for some conjunction of literals C(x) with $x$ free and  $(\neg \varphi(x) \wedge C(x)) \cftual \delta$ but $(\varphi(x) \wedge C(x)) \cftual \neg \delta$.  

\section{IB in hierarchical Bayesian learning}\label{sec:higherorder}

It's not unreasonable to rule out new evidence from unreliable testimony, {\em provided} the assignment of one's evaluation hypotheses to the testimony is reasonable.  But nothing in our discussion above forces the evaluation hypotheses to be be reasonable.  %, if the only alternative hypotheses that $\fh$ considers given his background beliefs in our hypothetical case make it that the confirming evidence for a hypothesis is completely discounted.  
Without any constraints, $\fh$'s evaluation hypotheses may rule out evidence that is completely grounded in reality and comes from testimony that an ideal rational agent would trust.  %Given the nature of some background beliefs, whole types of events can be and are discounted by many agents (take the case for example of climate denial).  %That is, $\fh$'s evaluation hypotheses may rule out any amount of good evidence or facts about the world.

To solve this problem, we need to correct the background beliefs ${\mathcal B}$.  Ideally, a rational agent should control for the biases in testimony by consulting several different bodies of testimony.  However, ${\mathcal B}$ cannot be corrected itself by evidence, % $x$, 
because that evidence is already discounted if it conflicts with ${\mathcal B}$.  Very clearly, background beliefs  can be a source of bad epistemic biases, and they can prevent straightforward corrections to improve one's beliefs as Bayesian learning would have us do.

Hierarchical Bayesian models were designed to address this problem \citep{gelman:etal:2013}.  In hierarchical Bayesian models, a Bayesian learning model like the one we have discussed in Section \ref{sec:firstorder} has certain parameters; the one parameter we have is our evaluation hypotheses providing the reliability of testimony.  At a second level of the hierarchy, we could have a Bayesian learning model concerning evaluation hypotheses, in which we could detail factors that would allow us to estimate reliably the accuracy of an evaluation hypothesis.  Abstractly, we would have evaluation hypotheses about evaluation hypotheses that would discuss factors like the consistency or the predictive accuracy of a testimony source, or the extent to which testimony from other sources agrees with its content.  One could also require a longer or more thorough exploration of the data about the phenomenon before the agent's restricting himself to a small subset for exploitation (once again an application of the work in \cite{cesa-bianchi:lugosi:2006}). All of these ideas and more have been proposed.

Simply requiring evaluation hypotheses that obey exogenous constraints, however, begs the question of why $\fh$ should accept them.  In fact, the interdependence of testimony, new information and background beliefs can make the resort to higher order parameters to resolve IB a failure % This is because testimony, especially {\color{violet} argumentatively complete testimony *give concrete example*}, may reinforce beliefs that preclude learning or even considering anything contrary to the testimony.  %For this to be the case, the testimony $t$ has to attack as groundless any body of testimony $t'$ that is inconsistent with it. 
because a body of dynamic testimony $T$, when directed by a conversational agent for the purposes of persuading and keeping his audience, can react to and attack not only a conflicting body of testimony $T'$ but also sources supporting it.  This behavior provides arguments for or against not only first order evaluation hypotheses, as we've seen with the notion of consistency, but also for higher order functions and in fact sequences of evaluation hypotheses. %If $T$ attacks $T'$ as untrustworthy, this implies that an adequate evaluation hypothesis $s$ according to $T$ should be such that $h(T')$ should be low if not $0$.  
%The purpose of an author behind such a body of testimony is to persuade and to capture the beliefs of his readers. 

To formalize this picture, we assume a hierarchy of sets of evaluation hypotheses where, $$h^{n+1}: h^n \rightarrow [0,1], \mbox{ for } h^{n+1} \in {\mathcal H}^{n+1}.$$  
\noindent
Higher order parameters at level $n+1$ are related to probabilities to evaluation hypotheses at level $n$ via a notion of {\em rationality}.  % We lay out the consequences of an argumentatively complete body of evidence in Proposition \ref{levels}, but first we define the notion of coherence for a set of sets of evaluation hypotheses, in which a set of evaluation hypotheses $h^{n+1}$ pronounces on the reliability of the elements of the set $h^n$. 
\begin{definition} \label{rational}
A set of sets of evaluation hypotheses ${\mathcal H} = \{\mathcal{H}^1, \mathcal{H}^2, ..., \mathcal{H}^n \}$ is {\sf rational} iff for all $m < n$, $h^{m}_k \in {\mathcal H}^m$, $P(h^{m}_k) = \lambda \sum_{h_j \in {\mathcal H}^{m+1}} P(h_j^{m+1}).P(h^m_k | h_j^{m+1})$ for some normalizing factor $\lambda$.
\end{definition}

\vspace{0.5mm}
\noindent 
A rational set of sets of evaluation hypotheses is thus one in which the probability of the evaluation hypotheses at one level reflects what the higher levels say about it.  Henceforth, we assume that agents' sets of evaluation hypotheses are rational.  

Given rational ${\mathcal H} = \{\mathcal{H}^1, \mathcal{H}^2, ..., \mathcal{H}^n \}$, we define a ${\mathcal H}^n$ sequence $\sigma \in \prod_{i=1}^n \mathcal{H}^i$ of consistent evaluation hypotheses to {\sf support} $T$ ($\sigma \csmodels T$) (or that make $T$ {\sf potentially trustworthy}---$\sigma \models T$) iff the ${\mathcal H}^1$ element $h^1_\sigma$ of $\sigma$ is such that $P(T_j|h^1_\sigma) $ is eventually monotone increasing with respect to $j$ (converges to $1$ as $j \rightarrow \infty$)  and every element of $\sigma$ has non-0 probability given ${\mathcal H}$.  Conversely, we say that $T \csmodels \sigma$ iff for each element $h^i_\sigma$ of $\sigma$ $P(h^i_\sigma | T_j)$ is eventually monotone increasing for all stages $T_j$. We note that $\sigma \csmodels T \rightarrow T \csmodels 
\sigma$. 

Let $\sigma^2$ to be the subsequence of $\sigma$ such that $\sigma^2 = \sigma\upharpoonright(\prod_{i = 2}^n \mathcal{H}^i).$ For $h^1 \in \mathcal{H}^1, \sigma^2(h^1)$ signifies the support $h_1$ receives from the higher order functions in $\sigma$ via Definition \ref{rational}.  
\begin{definition}
Given ${\mathcal H} = \{\mathcal{H}^1, \mathcal{H}^2, ..., \mathcal{H}^n  \}$, we say that an  ${\mathcal H}^n$ sequence $\sigma$ {\sf undercuts} $T$ iff for any $h^1 \in \mathcal{H}^1$ if $ P(T|h^1)> .5, \sigma^2(h^1) \leq  1- P(T|h^1)$. 
%$P(T|h) = 1 - \sigma^2(h)$.
%%$T \cup T'$ is inconsistent 
\end{definition}
\noindent

\begin{definition} $\phi$ {\sf disagrees} with $T'$ just in case $P(T|\phi) < P(T).$
\end{definition}
\begin{definition}
  $T$ {\sf attacks} $T'$ iff there is a ${\mathcal H} = \{\mathcal H^1, \mathcal H^2, ..., \mathcal H^n \}$ with: (i) ${\mathcal H}^m \mbox{ sequences } \sigma$ for $m <n$ such that: if $\sigma \models T$ $P(T'| h^1_\sigma) = 1 - P(T| h^1_\sigma)$ and $\sigma$ undercuts $T'$, and (ii) 
  for any ${\mathcal H}^m$ sequence $\sigma$, $m<n$ if $\sigma \csmodels T'$, $\exists h^{m+1} \in {\mathcal H}^{m+1}$ such that  $(P(h^{m+1} |T) > .5$ and  $h^{m+1}(\sigma) = 0)$.
\end{definition} 
\begin{definition}
%Let $T = \{T_1, T_2, ... , T_n, ...\}$ be a body of evidence.  
$T$ is {\sf argumentatively complete} iff:\\
 (i) $ (T' \models \phi \mbox{ and \em Disagree}(\phi, T)) \rightarrow \mbox{\em Attack}(T, T')$; (ii) If $T_n \not\models \phi$ but $P(T_n|\phi) \geq P(T_n)$, then $T_{n+1} \models \phi$.  (iii) for any T undercutting ${\mathcal H}^m$ sequence $\sigma$, $ \forall h^{m+1} \in {\mathcal H}^{m+1}$ such that  $(P(h^{m+1} |T) > .5, h^{m+1}(\sigma) = 0)$ (iv) $\exists {\mathcal H}^n \mbox{ sequence }\sigma $ such that $h_\sigma^1 \models T.$
 %there is a $h^{m+1}$ such that $h^{m+1}(\sigma) = 0$ and 
%for any $h \csmodels T$ there is an ${\mathcal H}^m$ sequence $\sigma_1$ such that  $T \csmodels h^{m+1}(\sigma_1^2(h))$. 
%\sigma_1 \csmodels T \ \wedge \ \sigma_2 \csmodels T') \\
%\rightarrow  \exists h_3 \ T \csmodels h_3 \wedge h_3(h_2) = 1 - P(h_1 | T)) ]

\end{definition}
\begin{proposition} \label{model-complete}
If $T$ is argumentatively complete, then $T$ supports a hypothesis $h$ that is PWMC for $T$.   
\end{proposition}
\noindent Assume that $T$ is argumentatively complete.  Then $\exists {\mathcal H}^n \mbox{ sequence }\sigma $ such that $h_\sigma^1 \models T$.   Now assume $T_n\not\models \phi$ for some $\phi$ for all stages $n$.  But then $P(T_n| \phi) < P(T_n)$ for each stage $T_n$ of $T$.  But then $T$ and $\phi$ disagree and so $T$ attacks $\phi$.  By the definition of attack,  $P(\phi | h^1_{\sigma_1}) = 1- P(T | h^1_{\sigma_1})$.  So $h_\sigma^1$ is PWMC. $\Box$

%Given Proposition \ref{model-complete}, an argumentatively complete $T$ supplies a complete view on evidence. Thus it is natural that when agents update $T$ supporting distributions ${\mathcal L}_\sigma$ on $T_i$, this shifts ${\mathcal L}_\sigma$ to distributions ${\mathcal L}_{h^1} | T_{i+1}$, where the first element $h^1$ of $\sigma$ is such that $(h^1 | T_{i+1})(T') = 1 - (h^1 | T_{i})(T_i)$ for all $i$ and for all $T' \neq T$.  Conditionalizing on argumentatively complete $T$ thus makes evaluation hypotheses dogmatic on $T$.  

%An argumentatively complete body of evidence always proposes a counter evaluation hypothesis to one that attacks its credibility, and thus resembles an acceptable argument in \cite{dung:1995}. 

  %we had before relative to S but that doesn't make sense

\begin{proposition} \label{levels} Let $T$ be argumentatively complete with a rational set of evaluation hypotheses ${\mathcal H}$ with $\sum_{h^1 \in {\cal H}^1}P(h^1) \neq 0$ and probabilities %of elements of $h^1$ 
updated on $T$.  
\begin{equation*}
\mbox{\em As } n \rightarrow \infty, \ {\sf P}_n(T) \rightarrow 1. \tag{1}
\end{equation*}
In addition suppose there is a $T' \not\subseteq T$.
\begin{equation*}
\mbox{\em As } n \rightarrow \infty, \ {\sf P}_n(T') \rightarrow 0. \tag{2}
\end{equation*}

%Suppose further that  an $h^1_1 \in {\mathcal H}^1$ such that $h^1_1 \models T$ and $P(h^1_1 | T_i) > .5$ in ${\mathcal H}$ for all $i$.%Suppose there is a $T'$ such that $T \cup T'$ is inconsistent.  %further that there are $h^1_1, h^1_2 \in {\mathcal H}^1$ such that $h^1_1 (T_1) = 1$ and $h^1_1(T'_1) = 0$,  and $h^1_2 (T_1) = 0$ and $h^1_2(T'_1) = 1$.  
%Then, for any sequence $\sigma \in \prod_1^k h_i$ for ${\mathcal H} = \{h^1, h^2, ..., h^n \}$ with elements $h^m \in {\mathcal H}^m, h^{m-1} \in {\mathcal H}^{m-1}$ for each $m < k$ such that $h^{m}(h^{m-1}_) > 0, ...$ and element $h^2$ such that $h^2(h^1_1) \neq 1$,  as $n \rightarrow \infty, \ {\sf P}_n(\sigma) \rightarrow 0$%, for all $k \geq 2$ and $\leq m$. % where $h^1(T) < 1$.% s
%$h^{n+1}_2 \in {\mathcal H}^{n+1}$ such that $\forall s \in {\mathcal H}^n, h^{n+1}_1(s) = 0$ iff $h^{n}_2(s) = 1$.
\end{proposition}  
\noindent
 %By Proposition \ref{codependence} $lim_{n \rightarrow \infty} {\sf P}_n(h^1_1) = 1$, and by Proposition \ref{discount} ${\sf P}_n(T) \rightarrow 1$.  
% If we can show that for argumentatively complete $T$ that $h^1_1$ must have non 0 probability relative to ${\mathcal H}$, and if we can show that any first order evaluation hypotheses $s$ such that not $s\csmodels T$ must have either 0 probability or be such that their probabilty decreases monotonically as we update with stages of $T$, then by 
We first show (1).  Since $T$ is argumentatively complete, $\exists \sigma \in {\mathcal H}$ such that $h^1_\sigma \models T$.  We need to show that for some such $h^1_\sigma$, $P(h^1_\sigma) \neq 0$ relative to ${\mathcal H}$.  
Suppose that $P(h^1_\sigma) = 0$, for all $h^1_\sigma$ such that $h^1_\sigma \models T$.  By rationality, for each such $h^1_\sigma$, $P(h^1_\sigma) =  \lambda \sum_{h_j \in {\mathcal H}^{2}} P(h_j^{2}).P(h^1_\sigma | h_j^{2}) = 0$.  Thus, all the non-0 probability mass of ${\cal H}$ falls on $T$ undercutting sequences $\sigma_i$. 
%there is a $T$ undercutting sequence $\sigma_1$ such that as ${\sf P}_n(T|h^1_\sigma) \rightarrow 1$, ${\sf P}_n(\sigma^2_1 |h^1_\sigma) \rightarrow 0$.  The non $0$ probability mass relative to ${\mathcal H}$ goes on $h^1_{\sigma_i}$ for $T$ undercutting $\sigma_i$.  
But for each such $T$ undercutting $\sigma_i$ of length $m$, since $T$ is argumentatively complete, there is an evaluation hypothesis $h^{m+1}$ supported by $T$ such that $P(\sigma_i |h^{m+1}) = 0$.  
%Since we update on $T$, the probability of $h^{m+1} > .5$ and by rationality all higher sequences must confer this probability on $h^{m+1}$.  
Since  ${\cal H}$ has only finitely many levels, at some level $k$ all T undercutting sequences $\sigma_j$ get $0$ probability.  This, together with the fact that $\sum_{h^1 \in {\cal H}^1}P(h^1) \neq 0$, contradicts the assumption that $P(h^1_\sigma) = 0$.
%, and $T \csmodels h^*(\sigma^*(h^1_\sigma)) $. By Coherence, $h^1_\sigma$ has non 0 probability.   In addition, by clause (ii) of the definition of an argumentatively complete $T$ and the definition of $T \csmodels h$, $P(h^1_\sigma|T_j)$ is monotonic increasing for all $j$.  Now consider some $h^1_2$ such that $P(T|h^1_2) < .5$.   Then $P(\neg T|h^1_2) > .5$.  $\neg T$ attacks $T$.  Since $T$ is argumentatively complete, for any sequence $\sigma \csmodels \neg T$, there is a $h^{m+1}$ such that $h^{m+1}(\sigma) = 0$ with $T \csmodels h^{m+1}(\sigma')$ for every $\sigma'$ such that $\sigma' \csmodels T$.  So $h^1_2$ has $0$ probability.  
Since $T$ is argumentatively complete, any sequence supporting any $h^1$ where $P(T|h^1) < P(T)$ will eventually get probability $0$; so $\sum_{\{h^1: P(T|h^1) \geq P(T)\}}P(h^1) =  \sum_{h^1 \in {\cal H}^1}P(h^1)$.  Moreover, as ${\sf P}_n$ gets updated, as $n \rightarrow \infty, \{h^1: {\sf P}_n(T|h^1) \geq {\sf P}_n(T)\} \rightarrow \{h^1: h^1 \models T \}$.  %So the non $0$ mass of {\cal H}$ must all go to sequences $\sigma$ where $\sigma \csmodels T$. 
%{\sf P}_n(h^1) \rightarrow 0$ as ${\sf P}_n(h^1_\sigma) \rightarrow 1$.  
The conditions on first order evaluation hypotheses in ${\mathcal H}$ of Proposition \ref{codependence} are now met. By Propositions \ref{codependence} and \ref{discount}, as ${n \rightarrow \infty}, \ {\sf P}_n(h^1_\sigma) \rightarrow 1, {\sf P}_n(h^1_i) \rightarrow 0$ for $i\neq 1$.    By Proposition \ref{discount}, ${\sf P}_n(T) \rightarrow 1$. 

To show (2), by Proposition \ref{model-complete}, $h^1_\sigma$ is also PWMC for $T$. As ${n \rightarrow \infty}, since \ {\sf P}_n(h^1_\sigma) \rightarrow 1$, ${\sf P}_n(T') \rightarrow 0.$ $\Box$

 \begin{proposition} \label{unlearning1}
Suppose $T$ is argumentatively complete.  Let $\fh$ be a hierarchical Bayesian learner whose evaluation hypotheses are rational and are updated on $T$. If $T' \subsetneq T$ such confirms a hypothesis $h$ that $T$ does not, then $\fh$ is incapable of learning $h$.  \end{proposition}
\noindent %By Proposition \ref{levels}, $\fh$'s higher order parameters will all eventually support the first order evaluation hypotheses that make $T$ trustworthy.   
%So by Proposition \ref{discount}, 
%If $T' \models h$ and $T\not\models h$, then since $T$ is model complete, $T \models \neg h$ and so $T'$ is inconsistent with $T$.  If $T'$ merely supports $h$ but $T\neq T'$, then $T$'s support is dogmatic on $T$.  
Claim 2 of Proposition \ref{levels} shows that
${\sf P}_n(T') \rightarrow 0$.  Then apply Proposition \ref{unlearning}. $\Box$ % But also ; so then by Proposition \ref{discount} ${\sf P}_n(T') \rightarrow 0$, and the result follows.

Argumentatively complete testimony thus collapses the case of higher order Bayesian frameworks to our first order setting.  What is troubling about IB is that our learner $\fh$ may hold onto an argumentatively complete $T$ regardless of how inadequate it is in the eyes of others or standard epistemic criteria; an argumentatively complete theory will always eventually find a reply to any attack or any doubt $\fh$ might acquire.

Argumentatively complete testimony isn't just an abstract concept; many social media and news sites already  approximate this condition. Outlets like {\em NewsMax} or {\em One Amercan News Network} that have a particular political bias will attack the credibility of stories from other bodies of testimony that have gone against a narrative they were and are promoting; darker conspiracy spinning websites like those promoting QAnon will attack arguments against their theories once they become aware of them.\footnote{See Stuart A. Thompson,
``Three Weeks Inside a Pro-Trump QAnon Chat Room'' {\em NY Times}, Jan 26, 2021).} In anecdotal support of our claims, consider Michelle Goldberg's ``It's Marjorie Taylor Greene's Party Now" {\em New York Times}, 2/2/2021) description of a group in IB: ``American conservatism — particularly its evangelical strain — has fostered derangement in its ranks for decades, insisting that no source of information outside its own self-reinforcing ideological bubble is trustworthy.''

A crucial component of argumentatively complete testimony $T$ is that it promotes evaluation hypotheses that both make $T$ eventually trustworthy but also PWMC for $T$.  Sources like {\em the New York Times} embody this in their slogan {\em all the news that's fit to print}, but there's a commercial reason for this outcome; news sites and social media are out to capture market share and so they naturally promote themselves as accurate and complete at least in a certain domain. The nature of contemporary testimony leads agents naturally to a situation where IB occurs.

How general are the results in Propositions \ref{levels} and \ref{unlearning1}?  \cite{wolpert:2018} argues that PAC, Statistical Physics Framework, VC, and supervised Bayesian learning are four different instantiations of his extended Bayesian formalism, which we use.  Thus our results should hold for other frameworks.

 %First thought, once faced with an argumentatively complete $T$ be suspicious, try to find other testimony.  
 %Voting/ observe effects of $T$ in the world if possible.  But formally this means in the game. we need it to be decomposition insensitive.  So that $E$ can control the other nodes, in particular ACCEPT no?
\section{Comparisons to Prior Work}

 IB is an epistemological bias that is clearly related to confirmation bias \citep{lord:etal:1979,nickerson:1998,oswald:grosjean:2004},  in which agents interpret new evidence in a way that confirms their beliefs, and to the framing biases of \cite{tversky:kahneman:1975,tversky:kahneman:1985}.  People tend to see in the evidence what they believe.  These forms of bias, however, concern how beliefs and bias influence interpretation, painting only part of the picture of IB (see also \cite{JOLLI}).  Further, unlike much of the psychological literature which finds epistemologically exogenous justifications for this bias \citep{dardenne:leyens:1995}, we show how IB  is a natural outcome of Bayesian updating, rational resource management and the belief interpretation co-dependence.   %in  the dynamic process of bias hardening that results from a learner performing a Bayesian update on a given body of testimony subsequently updating her beliefs based on a given interpretation of data conveyed by a body of testimony $T$ and  on how this effects her ability to learn. W
% IB is a kind of iterated confirmation bias brought on by the natural co-dependence of beliefs and interpretation of evidence. But,  we have shown how  In addition, as far as we know, this phenomenon has not been studied with rigorous techniques beforehand.  

IB  is a concrete application of the work on bandits in, determining optimal allocation of resources to the exploration and exploitation of sources \cite{whittle:1980,lai:robbins:1985,banks:sundaram:1994,burnetas:katehakis:1997,auer:etal:2002,cesa-bianchi:lugosi:2006,garivier:cappe:2011}.  It is also  related to work on generalization in machine learning. Epistemic biases affect generalization and learning capacity in ways that are still not fully understood \citep{lampinen:vehtari:2001,zhang:etal:2016,kawaguchi:etal:2017,neyshabur:etal:2017}. \cite{zhang:etal:2016} show that standard techniques in machine learning  for promoting good epistemic biases and generalization---training error minimization, regularization techniques like weight decay or dropout, or complexity measures used to minimize generalization error (the difference between training error and test error)---do not necessarily lead to good generalization and test performance.  Argumentatively complete testimony $T$ incorporates an adversarial attack mechanism against any good epistemic practices that might discount $T$.  It's this mechanism that guarantees IB.  % standard techniques impotent in improving learning performance. % At present we do not kno  w how IB relates to other techniques.

The argumentation literature \citep{amgoud:demolombe:2014,dung:1995} is also relevant to IB. If testimony $T$ is \textit{argumentatively complete}, then $T$ always provides a counterargument to an attack against $T$--much like an acceptable argument in \cite{dung:1995}.  In addition, however, an argumentatively complete $T$ also supports higher order evaluation hypotheses that support hypotheses that support $T$.  There are also important connections to the literature on trust \citep{castelfranchi:falcone:2010}; in our set up learning agents trust certain sources over others, and our higher order setting invokes a hierarchy of reasons.  Nevertheless, the argumentation and trust-based work of which we are aware is complementary to our approach.  An argumentation framework takes a possibly inconsistent belief base and imposes a static constraint on inference in such a setting.  Similarly, trust is typically modeled in some sort of static modal framework.  By contrast, ME learning games and the whole Bayesian framework are dynamic, with beliefs evolving under evidence  and game strategies evolving under agent interaction.  It is this dynamic evolution that is crucial to our approach and, we think, to modeling agents and learning. In sum, we are not looking at the problem of consistency, but rather the problems of entrenchment and bias.

\section{The complexity of IB}\label{sec:games}

IB is a result about learning.  IB is a suboptimal but natural outcome of the way contemporary bodies of evidence are set up and how humans interpret them. Given our set up, everything turns on what body of evidence on which to update and with which evidential hypotheses.

If IB is suboptimal, its effects are still more worrisome, because agents in the grip of IB are often unwilling or incapable of changing their beliefs so as to be able to learn.  Of course, our learner might just be happy with $T$; perhaps he needs no more accurate or more truthful body of testimony.  He may not be interested in learning anything beyond what $T$ presents him with.  In this section, however, we assume a learner who might be interested in learning but has difficulting escaping his IB prison.    We assume a rational learner $\fh$ who updates according to his evidential hypotheses; so if he has an evidential hypothesis that confers a high probability on some $T$, he will update on $T$.  We've seen that $\fh$ can get IB when he unduly restricts the bodies of evidence which serve as the basis of update or when he attends to an argumentatively complete testimony.  So key to removing IB is to get $\fh$ to change his hypotheses and consider other evidence that that to which he is wedded.

Anecdotally, we have a lot of evidence that IB is hard to escape\footnote{See Thompson, cited in note 2.} In general, however, we lack a precise analysis of its difficulty.   In this section, we introduce a game theoretic method that shows IB is not only hard to defeat but it can even be hard to detect (leading to self-deception).  We will see that the choice of epistemic paradigms is important.

%In Section \ref{sec:bckgnd} we gave a general definition of bias using counterfactuals.  In principle we can discover epistemic bias by considering the counterfactuals that describe $\hat{f}$'s behavior.  %But finding such counterfactuals  exponentially difficult \cite{marques-silva:etal:2019,ignatiev:etal:2020}.  In addition, we will see that the complexity of determining IB is complicated in the higher order case; 
 %An exhaustive search of the learning model is generally not feasible, especially in the higher order case, as our results below indicate.  Thus, we take a different path and introduce a game theoretic analysis with  which we will subsequently use to gauge the feasibility of escaping from or avoiding IB.  

 To motivate our approach, consider how an actual conversation might go between our learner $\fh$ in the grip of IB and a person $E$ who wants to correct his problem.  $E$ might question $\fh$'s reasons for believing some proposition $\phi$; she might try getting $\fh$ to consider different bodies of evidence $T'$ that might disconfirm $\phi$.  $\fh$ might accept $T'$ or he might argue against it---by providing, for example, reasons why $T'$ is not trustworthy or why the arguments supporting $T'$ are faulty.  $E$ might attack those arguments or provide new evaluation hypotheses for consideration.  Our ME games formalize this interaction.

%To formalize this interaction, we introduce a kind of message exchange game \citep{JPL,JOLLI}, %{\sf ME gam}e $\game = (\vocab^
%\infty,\win_0, \win_1)$ consists of a potentially infinite sequence of messages between two players $0$ and $1$, with winning conditions set by a Jury.   To define the messages, we need a vocabulary $V$ of moves or actions; these must be semantically interpretable formulas that represent either linguistic or other contents.  The intuitive idea behind an ME game is that a conversation proceeds in {\sf turns} where in each turn one of the players ``speaks'' or plays a string of elements from $V$. In addition, ME games keep track of ``who says what''. To model this, each player $i$ is assigned a copy $V_i = V\times\{i\} $ of the vocabulary $V$.   Conversations correspond to {\sf plays} of ME games that are the union of finite or infinite sequences in $\vocab$, denoted as $\vocab^
%\infty$.   We denote plays via $\play$.  A player's winning condition in an ME game is a subset of the set of all possible sequences of plays $\vocab^\infty$, denoted as $\Play$.  The Jury evaluates whether the conversational moves made by players or conversationalists are successful or not.  In the case our ME game is win-lose, an ME game $\game$ is specified by $\game = (\vocab^
%\infty,\win_0)$. %The Jury can also impose other constraints on plays.
In an {\sf ME learning game} $\game = (\vocab^\infty,\win)$, the two players, our investigator $E$ and our Bayesian learner $\hat{f}$, construct a larger "conversation" by consecutively playing finite strings from the vocabulary $V_0$ and $V_1$ respectively.  $\win$ specifies the winning condition of $E$.
%{\color {magenta} motivate this section with a game of conversational replies on Twitter, Lou Dobbs or Sean Hannity}
%players: our Bayesian learner $\hat{f}$ with body of evidence $T$, $E$ a teacher.
The vocabularies $V_i$ of an ME learning game $\game$ consists of sequences of evaluation hypotheses (with some abuse of notation, we'll take a single $h^n_j$ to be a one place sequence) and a predicate ACCEPT.  ACCEPT means that $\hat{f}$ accepts the last suggestion by $E$ and confers upon it a non zero probability mass. Our ME learning games are subject to several constraints. 
\begin{enumerate} 
\item [A.] {\em Knowledge first} \citep{williamson:2002}: this is a constraint from formal epistemology; $\hat{f}$ only adds a sequence $\sigma$ to $\prod^n_1 \mathcal{H}^i_{\fh}$ for $\mathcal{H}^i_{\fh} \in {\mathcal H}_{\hat{f}}$ % = \{\mathcal{H}^1_{\fh}, \ldots \mathcal{H}^n_{\fh}, \mathcal{H}^{n+1}_{\fh}...\}$ 
if he has no argument that attacks $\sigma$---in other words no evaluation hypothesis $h^{n+1} \in {\mathcal H}^{n+1}_{\hat{f}}$ such that $h^{n+1}(\sigma) = 0$. 
\item [B.] The Jury in an ME learning game is epistemologically competent; i.e. it sanctions only evaluation hypotheses that advance learning.
\item [C.] $E$ may only add sequences of evaluation hypotheses sanctioned by the Jury.  We assume this to be a finite set ${\mathcal H}_J$.  
\item [D.] Both players  must only propose consistent and rational sequences.
\item [E.] $\hat{f}$ has learned from some body of evidence $T$, which is common knowledge. 
\item [F.] $\hat{f}$ may only refuse a proposal of $E$ in the higher order setting, if he has a reason to do so.
\end{enumerate}
%We need to say more about epistemological competence. An epistemologically competent Jury sanctions evaluation hypotheses th

We define a sequence $\sigma \in \prod^n_1 {\mathcal H}^i$ to be {\sf positive} if for each element $h^{m+1}$ and $h^{m}$ of $\sigma$ $h^{m+1}(h^m) >> 0$.  A sequence $\sigma$ {\sf nullifies} a sequence $\sigma_1$, if for all $m$ and for $h^m_1$ of $\sigma_1$, $h^{m+1}$ of $\sigma$ is such that $h^{m+1}(h_1^m) = 0$.  We can have two sequences each one nullifying the other.  This formally represents an $n$ round argument, with each round $j+1$ offering a counterargument to the argument of round $j$.  We will say that a hypothesis $h^1$ is $T$ {\em positive} if $h^1$ is positive and $P(T |h^1) = 1$

We now define the moves of a game $\game$, in which we suppose a body of evidence $T$ that $\fh$ has attended to and a body of evidence $T'$ inconsistent with $T$. $E$ plays first, then $\fh$ then replies.  The game ends if $\fh$ plays ACCEPT, which implies that he adds a hypothesis $h^1_*$  to ${\mathcal H}^1_{\hat{f}}$, with a non-0 probability mass and with with high $P(T'|h^1_*)$, where $T' \cup T$ is inconsistent.
 \begin{enumerate}
 %\item [(m1)] $E$ may ask $\hat{f}$ for the bodies of evidence that he has learned from. 
 \item [(m1)] $E$ proposes 
% \item $\hat{f}$ may then reject $T'$ or ACCEPT, in which case the game ends.
% \item If $\hat{f}$ rejects $T'$, $E$ then 
$T'$-positive $h^{1} \in {\mathcal H}^1_J$ to be added to ${\mathcal H}^{1}_{\hat{f}}$. %such that % for some $T'$, $T' \models e$, $T  \cup T'$ inconsistent and %for some $h$, $T \models h$ and $P(T|e) << \frac{1}{2}$, 
     %$h^1(T') = 1$ 
     %for some $T'$ inconsistent with $T$.  
%     \item [(m2)]   Suppose at round $k$ of $\play$ in $\game$ $E$ has proposed a positve $h^1_*$. At $k+1$ $\hat{f}$ may refuse to add $h^1_*$ to $\mathcal{H}^1_{\hat{f}}$.  
     \item [(m2)]  Suppose at round $k \geq 1$ of $\play$ in $\game$ $E$ has proposed a $T'$ positive $h^1$. At $k+1$ 
$\hat{f}$ may play ACCEPT.
     \item [(m3)]  Suppose at round $k$ of $\play$ in $\game$ $E$ has proposed a a $T'$ positive $h^1$. At $k+1$ $\fh$ may play a nullifying $h^2 \in {\mathcal H}^1_{\hat{f}}$ such that $h^2(h^1) = 0$, if there exists such $h^2 \in {\mathcal H}^2_{\fh}$. 
      \item [(m4)] Suppose $E$ has proposed a positive sequence $\sigma$ of length $m$ and with $h^1_\sigma$ $T'$ positive at round $k$ of $\play$ in $\game$. At round $k+1$ $\hat{f}$ may respond  with sequence of length $m+1$ nullifying $\sigma$.
     \item [(m5)] Suppose at round $k$ of $\play$ of $\game$, $\hat{f}$ has proposed an m-length sequence $\sigma$ nullifying a positive $\sigma_*$ proposed by $E$, with $T'$ positive $h^1_{\sigma_*}$.  $E$ may respond at round $k+1$ of $\play$ with a positive $m+1$ length sequence $h_*^{m+1}.\sigma_*$, with $h^{m+1}_1(h^m) \neq 1$ for $h^m$ in $\sigma$. 
%    \item [(m5)] If $\hat{f}$ proposes an appropriate $h^2$, $E$ then may propose a positive sequence $\langle h^3_*, h^2_*, h^1_* \rangle$, where $h^3_*(h^2) \neq 1$, %$h^3_*(h^2_*) >> 0$, $h^2_*(h^1_*) >> 0$, 
  %  and $h^1_*(T') = 1$.   
    \item [(m6)]  Suppose   at round $k$ of $\play$ in $\game$, $E$ has proposed a positive sequence $\sigma$ of length $m$ and with $h^1_\sigma$ $T'$ positive.  At round $k+1$ $\fh$ may play ACCEPT, which implies that he adds $\sigma$ to $\prod^n_1 \mathcal{H}^i_{\fh}$.  
%    \item [(m7)] Suppose   at round $k$ of $\play$ in $\game$, $E$ has proposed an appropriate positive sequence $\sigma$.  At round $k+1$ $\hat{f}$ may respond with a refusal, with $E$ offering another positive sequence of the same length as $\sigma$ with play as in (m2).
   
%\langle h^4,h^3,h^2 \rangle \mbox{ with } h^i \in {\mathcal H}^i_{\hat{f}}$, such that $h^3(h^2_*) = 0$ and $h^4(h^3_*) = 0$, if there exist such  $h^2,h^3,h^4 \in \bigcup {\mathcal H}_{\fh}$.  
   % \item [(m9)] At each stage $n > 3$, $E$ may play a positive sequence of length $n+1$. $\hat{f}$ responds either with a sequence of length $h_{n+2}$ that patterns with the sequence in step (m7) or refuses or accepts and play proceeds as in (m6).
  
\end{enumerate} 
%$E$ may then propose another evaluation hypothesis $h^1_j$.  If $\hat{f}$ has refused all the functions proposed by $E$, the game ends. 

We note that if move (m6) occurs  $\hat{f}$, assigns $h^1_*$ and $T'$ a non-0 probability mass and updates with evidence $T'$. In which case the game ends. 
%{\color {magenta} Discussion of the game moves.  Why nullifying? because IB people don't want to look at those other sources...}

Suppose that in an ME learning game $\game$, $E$'s winning condition is simply to discover that $\hat{f}$ is interpretively blind, if he is. Call this condition $IB$.  We establish the complexity of $E$'s attempt to achieve $IB$.  The first order case with a finite ${\cal H}$ where the game is restricted to moves m1,m2,m3, is rather trivial.  
More interesting is the case of an ME learning game $\game =  (\vocab^
\infty,\win)$ with $\win = IB$  and in which $E$ and $\hat{f}$ play higher order evaluation hypotheses.    
\begin{proposition}  \label{blind-ho}
Suppose an ME learning game $\game = (\vocab^
\infty,\win)$ with $\win = IB$ in which $\hat{f}$ plays moves described in (m4)- (m7).  Then $\hat{f}$ is not interpretively blind iff play stops at some finite ordinal $n$.  %If play stops at some finite ordinal $n$ with $\hat{f}$ having refused all options from $E$ to consider alternative evaluation hypotheses $h^m_J$ for some $m < n$ and $m > 1$, $\hat{f}$ is higher order interpretively blind.  
\end{proposition} 
\noindent Suppose that in the play of $\game$, $\hat{f}$ accepts at some level $n$ to add the sequence of evaluation hypotheses proposed by $E$.  Then by the construction of the sequence and the requirement of coherence (constraint D), this confers upon some evaluation hypothesis $s*_1$ a non zero probability such that $P(T'|h^1_*) = 1$, where $T'$ is incompatible with the body of evidence $T$. By accepting, $\hat{f}$ will have an evaluation hypothesis $h^1_*$ with non zero probability such that $P(T'|h^1_*) = 1$, where $T'$ is incompatible with the body of evidence $T$, which $\hat{f}$ has proposed as a source of learning (constraint E).  Now when $\hat{f}$ updates his belief in $T$ he must do so with respect to $h^1_*$, and he must now update his confidence in his evaluation hypotheses with respect not only to $T$ but also $T'$.  In that case, $P(h^1_* | T_n, T'_n) \not\rightarrow 0$ and ${\sf P}_n(T') \not\rightarrow 0$.  As a result, $\hat{f}$ will be able to learn from $T'$, and so he is not interpretively blind with respect to $T$.

If there is no stopping point at any finite ordinal, then $E$'s is never able to get $\fh$ to accept a $T'$ positive hypothesis.  In which case, $\hat{f}$ continues to only update on $T$ and by Propositions \ref{levels} \ref{unlearning1}, $\hat{f}$ is interpretively blind. $\Box$

%\begin{cor} IB cannot be established through an exhaustive search of the set of counterfactuals describing $\fh$'s behavior.
%\end{cor}
%In Section \ref{sec:firstorder} we saw that a set of counterfactuals can define IB but that set is potentially infinite in the higher order case: it encodes E's winning strategy in an ME learning game.  %While an exhaustive search for true counterfactuals cannot establish higher order IB, 
%On the other hand, we can provide a finitary condition using ME games.  
%Technically, the winning condition of $E$ to determine whether $\hat{f}$ is subject to IB in an ME learning game $\game$ where higher order learning is invoked is rather simple.  $E$'s winning condition contains all those plays that never pass through the node ACCEPT (thinking of the moves in the game as nodes in a graph). Technically   %    Goals at the second level of the Borel Hierarchy and higher cannot be determined one way or the other by any finite prefix of an infinite game play (sequence of moves).  And thus, we predict that it is not straightforward to show that an agent is in the grip of IB, especially 

%Proposition \ref{blind-ho} allows us to assign a complexity class to IB in the Borel Hierarchy.    
%\begin{cor} Suppose that in an ME learning game $\game$ with $\win = IB$.  The complexity of $\win$ is $\Pi^1_1$ in the Borel Hiearchy.
%\end{cor}

Suppose $E$'s winning condition for an ME learning $\game$, is to get $\fh$ to accept a $T'$ positive evaluation hypothesis. Call this winning condition for $E$ ${\mathcal P}$ (for persuasion). %$\hat{f}$ to accept a sequence $\sigma$ of evaluation hypotheses $h^n_*.h^{n-1}_*.\ldots . h^1_*$, with $h^i_* \in  \bigcup \bigcup {\mathcal H}_{\hat{f}}$, whose level 1 element, $h^1_*$, has non 0 probability in virtue of $\sigma$ and the assumption of coherence of ${\mathcal H}_{\hat{f}}$ and renders trustworthy evidence not predicted by $T$.  By expanding the set of $\hat{f}$'s level 1 evaluation hypotheses, $\hat{f}$ will thus accord a non $0$ probability to $e$ that disconfirms some hypothesis of $T$.  
\begin{cor} \label{re} Suppose that in an ME learning game $\game$ with $\win = {\mathcal P}$.  The complexity of $\win$ is an R.E. set.  If $\win = IB$ then $\win$ is co-r.e.
\end{cor}
If ${\mathcal P}$ is the set of all finite plays, IB is its complement to a countable union of closed sets of plays in $\vocab^\omega$ that never pass through an ACCEPT move. $\Box$

\begin{proposition}
Suppose an ME learning game $\game$ with $\win = {\mathcal P}$ and $\hat{f}$  as described in Proposition \ref{levels}.    Then $E$ has no winning strategy in $\game$.  \label{nowin}
\end{proposition} 
 \noindent %$E$ wins $\game$ if she gets $\hat{f}$ to add a sequence of evaluation hypotheses in ${\mathcal H}_{\hat{f}}$ such that $h^1$ has non 0 probability.  
 Proposition \ref{unlearning1} implies $\hat{f}$'s evaluation hypotheses are updated on an argumentatively complete body of evidence $T$.  When implemented via an ME game $\game$, the sequence of evaluation hypotheses in Proposition \ref{levels} provide a winning strategy for $\hat{f}$.  Suppose $E$ proposes an $h^1$ supporting $e$ that is inconsistent with $T$.  Even if $E$ %exploits an argumentatively complete theory $T'$ to 
 generates a suitable sequence of higher order $T'$ positive evaluation hypotheses $h^1, h^2, h^3, \ldots$, given Constraint A above, $\hat{f}$ will only accept an evaluation hypothesis if he has no argument against it.  But as $T$ will eventually supply such an argument, $\fh$ can always counter $E$'s proposals.  So she has no winning strategy. $\Box$

\noindent Not only is IB computationally complex (Corollary \ref{re} shows it is not computable but $\Pi_1$),  Proposition \ref{nowin} shows formally that %one cannot simply resort to higher level constraints to get even an epistemically rational player out of IB; the “rot” imposed by IB goes ``all the way up.''   E
even if $E$ has rationally compelling arguments to show that $\hat{f}$ is better off (his payoff or reward is higher) in accepting her proposed sequence of evaluation hypotheses, $\hat{f}$ can rationally resort to $T$ to counter her argument.  Extracting someone from higher order IB is thus impossible by purely epistemic means.  There is {\em no way} of getting someone, even a rational agent, out of higher order IB by purely epistemic arguments, given our assumptions.
This pessimistic is borne out empirically: some people in the grip of right wing conspiracy theories in the US were dying of Covid19 in December of 2020 and January 2021 but continued to refuse to believe that it was that disease that was killing them---despite all the evidence and arguments they were given, they refused to let go of an obviously faulty but argumentatively complete $T$.
%{\color {magenta} should the following commented out remarks remain?} %By exploiting a consistent, argumentatively complete body of evidence $T$ against an opponent who attacks $T$, one can construct an ME play that is rhetorically rational, responsive to attacking arguments and consistent; as such, this play meets the necessary conditions for a winning play in an ME game according to proposition 4 of \cite{JPL}.  Nevertheless, we have seen that the learning theory that $T$ supports is clearly not a good one; it ignores facts and cannot learn properly.

Of course, people sometimes \textit{do} change their minds and do escape the grip of argumentatively complete theories, many times for epistemically exogenous reasons.\footnote{For instance, the satisfaction they derived from belonging to a particular community supported by a particular body of testimony might and does wane.}  But by challenging one of our assumptions, rational agents can of course also reject IB.  The weak link in our argument is assumption $A$, the "knowledge first" assumption.  Perhaps $\fh$ should accept evaluation hypotheses even if $T$ attacks them.  More likely, $\fh$ should not accept all attacks equally; he should be skeptical of any body of evidence $T$ that promotes  PWMC for $T$ and $T$ eventually trustworthy evaluation hypotheses while attacking any point of view at variance with it.  
%If such principles can be instilled prior to $fh$'s settling on a particular body of argumentatively complete testimony, then IB will never arise.

We now explore the  play between $E$ and $\fh$ in an ME learning game $\game$ where $\win = {\mathcal P}$ before $\fh$ has accepted enough of the argumentatively complete $T$ to close off learning from alternative bodies of evidence.  Suppose $T$ is argumentatively complete but comes in stages; if $T'_i$ attacks $T_i$, then $T_{i+1}$ but not $T_i$ attacks $T'_i$.  That is, an argumentatively complete $T$ reacts to attacks but does not forsee all attacks in advance.  %The picture now is more complex than when we looked at stages in Proposition \ref{codependence}, since we now may revise evaluation hypothesis values in light of new evidence.  
Suppose a set of consistent first order evaluation hypotheses ${\mathcal H}^1 = \{h^1_1, h^1_2, ... \}$, with $P(h^1_1) = .6, P(h^1_2) = .4$, and $P(T_i |(h_1) = 1 = P(T'_i | h_2)$.  %If $T_1 \cup T'_1$ is consistent, we may suppose that $h^1_1 \models T'_1$ and $h^1_2 \models T_1$. 
Now suppose $T'_1 \cup T_1$ is inconsistent and $E$ proposes $h^1_2$ since $h^1_2 \models T'_i$.  Since the $h^1_i$ are consistent, $P(T_1 | h^1_2) = 0 = P(T'_2 | h^1_1)$.  At this point, $\fh$ could accept $E$'s proposal under constraint (A), $\game$ ends and $E$ wins.  $\fh$ will continue to update over stages $T$ and $T'$ with the marginal probabilities  $P(T_i) = .6$ and $P(T'_i) = .4$ remaining stationary.  

On the other hand, $\fh$ may decide to wait to see what the next stage $T_2$ of $T$ brings.  As $T$ is argumentatively complete, $T_2$ will attack $T'_2$, and add a nullifying $h^2 \in {\mathcal H^2}$ supported by $T_2$.  Should $\fh$ accept $h^2$, the probability of $h^1_2$ will go to $0$ in ${\mathcal H}$.  But now suppose we have a constraint, {\sf Discount},that discounts any nullifying sequence from $T$.  It would be unreasonable for $\fh$ to wipe out alternatives in the face of this level of uncertainty; at this stage, $P(T_2) = .6$ and $P(T'_2) = .4$.     Summarizing:

\begin{proposition}
Suppose an ME learning game $\game$ with constraint $A$ replaced by Discount and with $\win = {\mathcal P}$ and $\hat{f}$ as described in Proposition \ref{levels}. $E$ then has a winning strategy in $G$, and IB does not arise for $\fh$.
\end{proposition}

 %First thought, once faced with an argumentatively complete $T$ be suspicious, try to find other testimony.  
 %Voting/ observe effects of $T$ in the world if possible.  But formally this means in the game. we need it to be decomposition insensitive.  So that $E$ can control the other nodes, in particular ACCEPT no?

\section{Conclusions}

Interpretive blindness results from a dynamic, iterative  process whereby a learner's background beliefs and biases lead her to update her beliefs based on a body of testimony $T$, and then biases inherent in $T$  come back to reinforce her beliefs and her trust in $T$'s source(s), further biasing her towards these sources for future updates.  We have introduced and formally characterized IB.  We have shown that IB can prevent learning even in higher order Bayesian frameworks for learning from argumentatively complete testimony, despite the presence of constraints designed to promote good epistemic practices.  We also shown that IB is computationally complex as a co-r.e. set via a game theoretic analysis, and that an agent may rationally remain in IB in the face of epistemic arguments. Our game theoretic analysis can also be extended to cases where the agent falls out of IB but then is a recidivist and becomse a prisoner once more.  We leave that for future work.

Investigating IB alas is not just an academic enterprise. IB really does happen, with sometimes tragic or dangerous results. We think a careful formal analysis is urgent for society.  Finally, we note that while we have focused on IB as a problem for learning from testimony, the problem it raises for learning extends to any case in which we do not have unmediated access to ground truth and our data is ``theory laden'' \cite{hanson:1958}. 
\bibliographystyle{ACM-Reference-Format} 

%\bibliography{coling2020}
%Now moving back to language.  
%\bibliographystyle{splncs04}
%\begin{scriptsize}
\bibliography{erc2}
\end{document}